\def\paperID{0000}
\def\confName{CVPR}
\def\confYear{2024}

\def\paperTitle{Woodpecker: Hallucination Correction for\\ Multimodal Large Language Models}

\def\authorBlock{
   \textbf{Shukang Yin$^{1}$\thanks{Equal contribution.}\hspace{.4mm}, ~Chaoyou Fu$^{2*\ddagger}$\thanks{Project leader.}\hspace{.8mm}, ~Sirui Zhao$^{1*\ddagger}$, ~Tong Xu$^{1\ddagger}$, ~Hao Wang$^{1}$}\\
   \textbf{Dianbo Sui, ~Yunhang Shen$^{2}$, ~Ke Li$^{2}$, ~Xing Sun$^{2}$, ~Enhong Chen$^{1}$\thanks{Corresponding author.}\hspace{1.mm}}\\
$^1$School of Data Science, USTC \& State Key Laboratory of Cognitive Intelligence \\
$^2$Tencent YouTu Lab \\
\texttt{\{xjtupanda,sirui\}@mail.ustc.edu.cn}, ~\texttt{\{tongxu,cheneh\}@ustc.edu.cn} \\
\texttt{\{bradyfu24\}@gmail.com} \\
}

\newif\ifreview 
\newif\ifarxiv \newcommand{\arxiv}{\arxivtrue}
\newif\ifcamera 
\newif\ifrebuttal 

\arxiv 

\pdfoutput=1
\documentclass[10pt,twocolumn,letterpaper]{article}
\ifreview \usepackage[review]{cvpr} \fi
\ifarxiv \usepackage[pagenumbers]{cvpr} \fi
\ifrebuttal \usepackage[rebuttal]{cvpr} \fi
\ifcamera \usepackage{cvpr} \fi

\usepackage[utf8]{inputenc}
\usepackage{graphicx}
\usepackage{bbding, pifont}
\usepackage{amsmath}	
\usepackage{amssymb}	
\usepackage{booktabs}
\usepackage{times}
\usepackage{microtype}
\usepackage{epsfig}
\usepackage[table,xcdraw,dvipsnames]{xcolor}
\usepackage{caption}
\usepackage{float}
\usepackage{placeins}
\usepackage{color, colortbl}
\usepackage{stfloats}
\usepackage{enumitem}
\usepackage{tabularx}

\usepackage[normalem]{ulem}
\useunder{\uline}{\ul}{}

\usepackage{tikz}

\usepackage{xstring}
\usepackage{multirow}
\usepackage{xspace}
\usepackage{url}
\usepackage{subcaption}
\usepackage{xcolor}
\usepackage{tcolorbox}
\usepackage[hang,flushmargin]{footmisc}

\ifcamera \usepackage[accsupp]{axessibility} \fi

% ...

\def\ie{\emph{i.e}\onedot} 
 
\def\etc{\emph{etc}\onedot}

\ifarxiv  \fi

\newcommand{\R}[1]{{%
    \textbf{%
        \ifstrequal{#1}{1}{\textcolor{red}{R#1}}{%
        \ifstrequal{#1}{2}{\textcolor{blue}{R#1}}{%
        \ifstrequal{#1}{3}{\textcolor{magenta}{R#1}}{%
        \ifstrequal{#1}{4}{\textcolor{teal}{R#1}}{%
                           \textcolor{cyan}{R#1}%
        }}}}%
    }%
}}

\usepackage{xcolor,amsmath}
\usepackage[linesnumbered,ruled,vlined]{algorithm2e}
\DontPrintSemicolon

\usepackage{xcolor}
\definecolor{mygreen}{HTML}{3cb44b}

\SetKwComment{Comment}{\color{green!50!black}\# }{}

\SetKwProg{Function}{def}{:}{}

\SetKwProg{For}{for}{:}{}
\SetKwProg{If}{if}{:}{}
\newcommand{\VarSty}[1]{\textnormal{\ttfamily\color{blue!90!black}#1}\unskip}

\usepackage{xr-hyper}

\makeatletter
\newcommand*{\addFileDependency}[1]{
  \typeout{(#1)}
  \@addtofilelist{#1}
  \IfFileExists{#1}{}{\typeout{No file #1.}}
}

\makeatother

\definecolor{cvprblue}{rgb}{0.21,0.49,0.74}
\usepackage[pagebackref,breaklinks,colorlinks,citecolor=cvprblue]{hyperref}
\usepackage[capitalize]{cleveref}
\crefname{section}{Sec.}{Secs.}
\crefname{table}{Table}{Tables}
\crefname{figure}{Fig.}{Figs.}

\begin{document}
\title{\paperTitle}
\author{\authorBlock}
\maketitle

\begin{tikzpicture}[remember picture,overlay,shift={(current page.north west)}]
\node[anchor=north west,xshift=3.3cm,yshift=-3cm]{\scalebox{-1}[1]{\includegraphics[width=2.2cm]{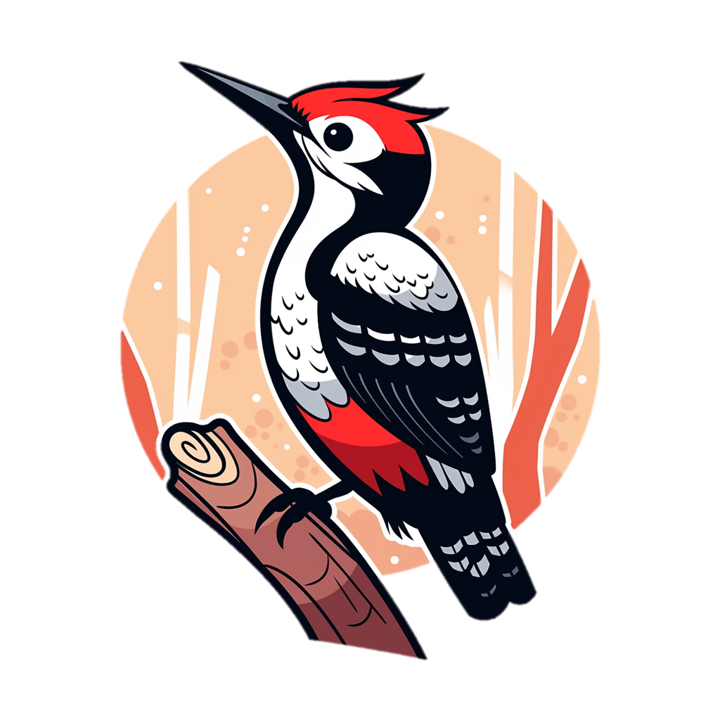}}};
\end{tikzpicture}

\begin{abstract}
Hallucination is a big shadow hanging over the rapidly evolving Multimodal Large Language Models (MLLMs), referring to the phenomenon that the generated text is inconsistent with the image content. In order to mitigate hallucinations, existing studies mainly resort to an instruction-tuning manner that requires retraining the models with specific data. In this paper, we pave a different way, introducing a training-free method named \textbf{Woodpecker}. Like a woodpecker heals trees, it picks out and corrects hallucinations from the generated text. Concretely, Woodpecker consists of five stages: key concept extraction, question formulation, visual knowledge validation, visual claim generation, and hallucination correction. Implemented in a post-remedy manner, Woodpecker can easily serve different MLLMs, while being interpretable by accessing intermediate outputs of the five stages. We evaluate Woodpecker both quantitatively and qualitatively and show the huge potential of this new paradigm. On the POPE benchmark, our method obtains a \textbf{30.66\%/24.33\%} improvement in accuracy over the baseline MiniGPT-4/mPLUG-Owl.
The source code is released at \url{https://github.com/BradyFU/Woodpecker}.
\end{abstract}

\section{Introduction}
\label{sec:intro}


\begin{figure}[t]
    \centering
    \includegraphics[width=0.92\columnwidth]{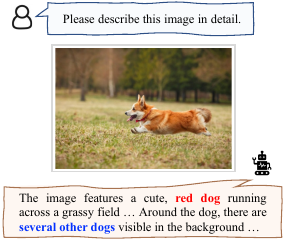}
    \caption{Illustration of hallucinations in MLLMs. Given an image, an MLLM outputs a corresponding response with both \textcolor{blue}{object-level} and \textcolor{red}{attribute-level} hallucinations. }
    \label{fig:motive}
\end{figure}

\begin{figure*}[!thb]
\centering
\includegraphics[width=0.97 \linewidth]{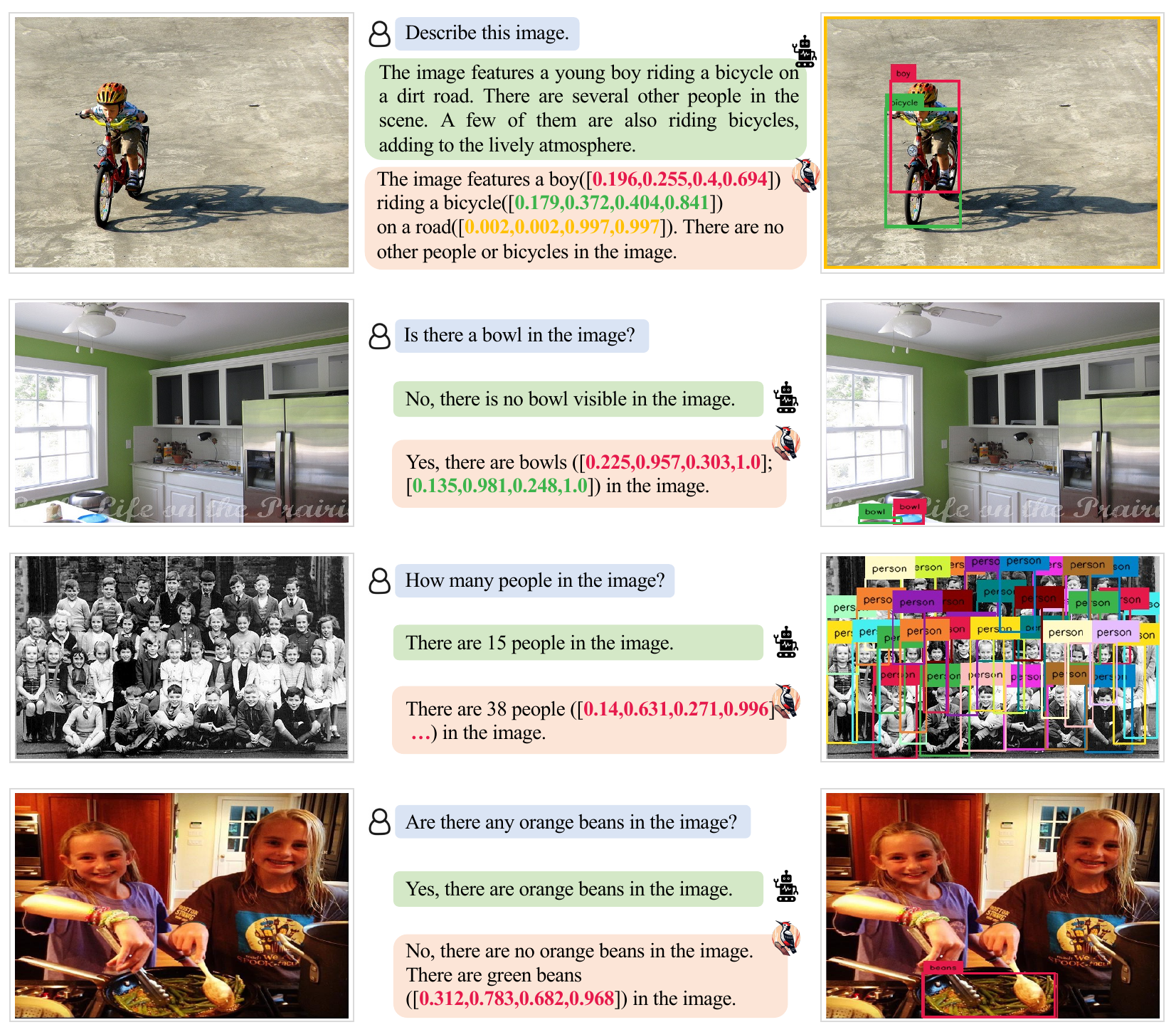}
\caption{
Examples of our framework for hallucination correction. Given a response of an MLLM, Woodpecker corrects the hallucinated parts and incorporates grounding information for ease of verification. 
}
\label{fig:teaser}
\end{figure*}

Multimodal Large Language Models (MLLMs)~\cite{yin2023survey} are now flourishing in the research community, working towards Artificial General Intelligence (AGI). By exploiting powerful Large Language Models (LLMs), researchers align foreign modalities like vision with language, and develop MLLMs with various exciting capabilities~\cite{liu2023visual,ye2023mplug,zhu2023minigpt,zhang2023transfer,Qwen-VL}, such as fully describe the contents of a given image.

However, as strong as these MLLMs are, they sometimes output descriptions that are inconsistent with the input image. It is called hallucination and has been found prevalent in MLLMs~\cite{liu2023mitigating}. As exemplified by~\cref{fig:motive}, the MLLM claims non-existent objects and fails to describe the attribute of the object in the image accurately, which are categorized by us as object-level and attribute-level hallucinations, respectively. 
It is obvious that these hallucinations are huge obstacles to the practical application of MLLMs.

In order to mitigate the hallucinations, existing works usually explore an instruction-tuning way~\cite{wang2023vigc,liu2023mitigating}. 
A common key observation is that MLLMs tend to hallucinate when generating longer text~\cite{liu2023mitigating}, which results in different problem-solving strategies. For example, LRV-Instruction~\cite{liu2023mitigating} takes an intuitive approach by limiting the text length of instruction data. 
As a consequence, the tuned model usually generates less hallucinated but also less detailed descriptions. VIGC~\cite{wang2023vigc} takes a multi-step generation scheme and iteratively updates the visual features with the textual context, which relieves hallucinations via sacrificing generative efficiency. Moreover, both of the two methods are instruction-tuning-based and thus are data- and computation-intensive.

To break the limitation, we take a different strategy that can directly correct the hallucinations without retraining. As illustrated in~\cref{fig:teaser}, given a text generated by MLLMs as well as the input image, our training-free framework Woodpecker corrects the text elaborately, and meanwhile, provides the corresponding evidence, \ie, the bounding boxes. It adds interpretability and reliability beyond the black-box MLLMs, providing convenient visual fact-checking.
Concretely, our framework performs correction after a thorough diagnosis, which incorporates a total of five stages: (1) \emph{Key concept extraction} identifies the main objects mentioned in the generated sentences; (2) \emph{Question formulation} asks questions around the extracted objects, such as their number and attributes; (3) \emph{Visual knowledge validation} answers the formulated questions via expert models. For example, a visual perception model can be used to determine the object number; (4) \emph{Visual claim generation} converts the above Question-Answer (QA) pairs into a visual knowledge base, which consists of the object-level and attribute-level claims about the input image; (5) \emph{Hallucination correction} modifies the hallucinations and adds the corresponding evidence under the guidance of the visual knowledge base. It is worth noting that each step in the pipeline is clear and transparent, which offers good interpretability.

We evaluate the effectiveness of our method through comprehensive quantitative and qualitative experiments on the POPE~\cite{li2023evaluating}, MME~\cite{fu2023mme}, and LLaVA-QA90~\cite{liu2023visual} datasets. 
The results and associated analyses indicate the superiority of this new paradigm. 
For instance, on the POPE benchmark, our method largely boosts the accuracy of the baseline MiniGPT-4~\cite{zhu2023minigpt}/mPLUG-Owl~\cite{ye2023mplug} from 54.67\%/62\% to 85.33\%/86.33\%.

In summary, the main contributions are as follows:

\begin{itemize}
\item We propose a training-free framework named Woodpecker to correct the hallucinations for MLLMs. To the best of our knowledge, we are the first to apply a corrective manner to tackle the visual hallucination problem.
    
\item Our framework is designed in a way that each step is clear and transparent, thus providing good interpretability.
    
\item We comprehensively evaluate the effectiveness of our method, and the large improvements demonstrate its great potential in hallucination correction.
\end{itemize}
\section{Related Work}
\label{sec:related}

\subsection{Hallucination in MLLM}
Recently, there has been increasing attention on the hallucination phenomenon of MLLMs. This is mainly because the issue directly affects the reliability of MLLMs. Current researches on the hallucination of MLLMs mainly focus on two aspects, \ie, the evaluation/detection~\cite{li2023evaluating,wang2023evaluation,gunjal2023detecting} and mitigation~\cite{liu2023mitigating,wang2023vigc,lu2023evaluation}. The previous line of work generally either trains a classification model to discriminate hallucination~\cite{gunjal2023detecting} or checks the output text against ground-truth answers to decide if the hallucination happens~\cite{li2023evaluating, wang2023evaluation}. 

For hallucination mitigation, previous works focus on optimizing the data collection process and the training scheme. LRV-Instruction~\cite{liu2023mitigating} composes negative instances to refrain from over-confidence. Moreover, the text length of Ground-Truth answers is strictly controlled, based on the observation that shorter responses are less likely to be hallucinated. Similarly, VIGC~\cite{wang2023vigc} takes an iterative process, where short answers are generated and concatenated each time. In this way, it tries to ensure accuracy without compromising details. 
While previous works try to develop MLLMs with fewer hallucinations, our main objective is to refine the responses of MLLMs by modifying the hallucinated parts. Specifically, we design a training-free framework that incorporates off-the-shelf models. This exempts the complexity of collecting instruction data and resource-intensive training. 
As a result, our framework can be easily integrated with various MLLMs, serving as a general plug-and-play module.

\subsection{Knowledge-augmented LLM}
Since LLMs are limited to the inherent knowledge gained from pretraining, various works have been dedicated to augmenting LLMs with external knowledge sourced from a pre-defined knowledge base~\cite{dinan2018wizard, petroni2020kilt, lewis2020retrieval, borgeaud2022improving} or the internet~\cite{shuster2022blenderbot, piktus2021web}. As a natural extension of this idea, recently, researchers have explored using knowledge as evidence to alleviate factual hallucinations in LLMs~\cite{huang2023zero,peng2023check}. Specifically, these works use relevant knowledge as background information to refine a possibly false input claim, resulting in a higher factuality of the response. Our methods share in common with the idea that we use information relevant to the given image to correct potentially wrong claims. 
However, it is non-trivial to transfer the idea to the vision-language field. 
This is because the language-only counterpart usually deals with text only and acquires relevant knowledge through retrieval, while it is inappropriate to do so for image-text pairs. Moreover, knowledge-augmented LLMs pay more attention to alleviating factual fallacies, while we lay more stress on mitigating visual hallucinations. 
Corresponding to the key differences, in this work, we devise a strategy to construct a structured visual knowledge base conditioned on the image and the query. We also explore how to address both object-level and attribute-level hallucinations in an organized way, as we will illustrate later.

\subsection{LLM-aided Visual Reasoning}
According to the taxonomy in the survey~\cite{yin2023survey}, our proposed framework is closely related to the LLM-Aided Visual Reasoning model~\cite{berrios2023towards,lai2023lisa,gong2023mindagent}. 
The main idea is that we can leverage the strong reasoning and instruction-following capabilities of LLMs to help fulfill vision or multimodal tasks. Typical roles that LLMs play include the task dispatcher~\cite{shen2023hugginggpt, yang2023gpt4tools, lu2023chameleon, gupta2023visual}, the reasoner~\cite{wu2023visual,zhu2023chatgpt,yang2023mm,you2023idealgpt}, or the language refiner~\cite{wang2023caption,zhang2023prompt,zhu2023pointclip,zeng2022socratic}. In this work, we utilize the strong reasoning and language proficiencies of LLMs to help the processes of key concept extraction, question formulation, and hallucination correction.

\section{Method}
\label{sec:method}

\begin{figure*}[!thb]
\centering
    \includegraphics[width=\linewidth]{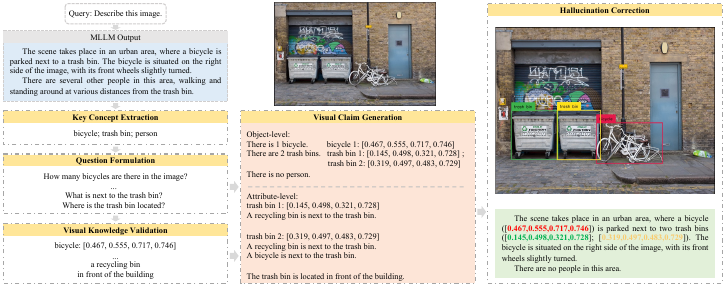}
    \caption{Framework of Woodpecker. Given an image and a query, an MLLM outputs the corresponding response. Through the four steps, including key concept extraction, question formulation, visual knowledge validation, and visual claim generation, we get a visual knowledge base specific to the image and the original response. In the final step, the hallucinations in the response are corrected with the bounding boxes as evidence.}
    \label{fig:framework}
\end{figure*}

Our objective is to diagnose and correct the hallucinations in the response generated by MLLMs. The key challenges lie in locating the hallucinations and determining the facts, which can be organized in a structured way for final correction. 
To this end, we break down the whole process into five subtasks: key concept extraction~(\cref{sec:method_1}), question formulation~(\cref{sec:method_2}), visual knowledge validation~(\cref{sec:method_3}), visual claim generation~(\cref{sec:method_4}), and hallucination correction~(\cref{sec:method_5}). We will illustrate each step in sequence later. 
An overview of our framework is depicted in~\cref{fig:framework}.

\subsection{Key Concept Extraction}
\label{sec:method_1}
Since descriptions usually revolve around key concepts, the first step is to extract them from the generated sentence. To this end, we identify the main objects mentioned in the sentence, which are the ones most likely to exit visual hallucinations.
For instance, given a sentence ``\texttt{The man is wearing a black hat.}'', the objects ``\texttt{man}'' and ``\texttt{hat}'' are extracted, and will serve as the center for diagnosis in the following steps.
In light of the strong generalization ability and rich world knowledge of LLMs, we prompt an LLM to fulfill this task.

The template for key concept extraction is listed in~\cref{ssec:prompt_entity}, which comprises a system message and a formatted prompt. The former sets up the basic context for the LLM, while the latter starts with some detailed descriptions of the task and some requirements, followed by several in-context examples and inputs. 
The in-context examples are provided so that the LLM could better understand the requirements in terms of the task.

\subsection{Question Formulation}
\label{sec:method_2}
After acquiring the key concepts, we ask a series of questions around them to make the hallucination diagnosis.
Our questions are directed at both object-level and attribute-level hallucinations.
For the former, we ask, ``\texttt{Is there any \{object\} in the image? How many are there?}'', where ``\texttt{\{object\}}'' is the key concept extracted earlier.
For the latter, various questions involving the attributes of objects can be formulated, such as ``\texttt{What is \{object\} doing?}'', ``\texttt{Is \{object$_1$\ on the right side of \{object$_2$\}?}'', and ``\texttt{What color is the \{object\}?}'', where ``\texttt{\{object$_1$\}}'' and ``\texttt{\{object$_2$\}}'' are different key concepts.

In fact, object-level questions can be directly validated through perceiving images, while attribute-level questions are much more diverse and dependent on the context. 
To facilitate such free-form formulation of questions, we prompt an LLM with some in-context examples so that meaningful questions are raised. 
The prompt is listed in~\cref{ssec:prompt_question}.

\subsection{Visual Knowledge Validation}
\label{sec:method_3}
This step is responsible for solving the above two types of questions. 
For the object-level questions, the crux is determining the existence and the count of a certain object.
In light of the strong perception capabilities of vision foundation models~\cite{kirillov2023segany, wang2023detecting,li2022grounded,singh2022flava,luddecke2022image}, we employ an open-set object detector as the solver~\cite{liu2023grounding}. 
For attribute-level questions, we apply a pre-trained VQA model~\cite{li2023blip} to answer the questions conditioned on the image. Compared with mainstream MLLMs, the VQA model tends to generate shorter answers but also with fewer hallucinations and thus can be a reasonable choice.

\subsection{Visual Claim Generation}
\label{sec:method_4}
After questions are raised and answered, we combine QA pairs into visual claims and organize them into a \textbf{visual knowledge base} for reference in the following step. The visual knowledge base is structured by:
\begin{itemize}
    \item Object-level claims: This part of the information mainly plays a role in mitigating object-level hallucinations. We include information about object counts of key concepts extracted from the sentences~(\cref{sec:method_1}). For existing objects, we add a claim as ``\texttt{There are \{counts\} \{name\}.}'', where ``\texttt{\{counts\}}'' and ``\texttt{\{name\}}'' are the counts and the name of a certain kind of object. We use a similar template, ``\texttt{There is no \{name\}}'', for nonexistent objects. The counting information comes from the open-set object detection in the previous step.

    \item Attribute-level claims: We include attribute information specific to each object in order to alleviate attribute-level hallucinations. Typical attributes include positions, colors, actions, \etc. For this part, we adopt a QA-to-Claim model~\cite{huang2023zero} to merge questions and answers into claims. 
    In order to cope with cases involving multiple objects or the relationship between the foreground objects and the background, more global information is needed. Thus, we also include claims that involve the interaction between different objects or the objects and the background, such as ``\texttt{The cat is lying next to the dog.}''.
\end{itemize}

\subsection{Hallucination Correction}
\label{sec:method_5}
Guided by the visual claims, an LLM can act as a corrector and modify the hallucinations in the generated responses.
Specifically, after combining the visual knowledge base with the original responses into a prompt, we instruct an LLM to correct the responses and output the refined ones. 
For better interpretability, we explicitly instruct the LLM to attach bounding boxes right behind expressions when referring to objects. 
This design facilitates the correspondence between the mentioned entities in the responses and object instances in the image, which provides convenient access to check the reliability of the output. 
The prompt template for correction is included in~\cref{ssec:prompt_correction}

\section{Experiment}
\label{sec:experiment}

\subsection{Experimental Settings}
\paragraph{Dataset.}
\textbf{POPE}~\cite{li2023evaluating} is dedicated to evaluating hallucinations of MLLMs.
It contains the settings of random, popular, and adversarial sampling, which mainly differ in the way negative samples are constructed.
For the random setting, the objects not presented in the image are sampled randomly, while for the popular setting, non-existent objects are sampled from a pool of objects with the highest frequencies. For the adversarial setting, objects that most frequently co-occur but do not exist in the image are sampled. 

In terms of the sampling setting, we sample 50 images and build 6 questions for each image. The ratio between positive and negative samples is balanced, namely 50\% vs 50\%.
This setup transforms object annotations into a series of ``Yes-or-No'' questions and focuses on evaluating the object-level hallucination, and more specifically, the \emph{existence} aspect. Thereby, MLLMs are prompted to answer if an object exists in the image or not. Accordingly, evaluation metrics include accuracy, precision, recall, and f1-score. 

\textbf{MME}~\cite{fu2023mme} is a comprehensive benchmark designed to evaluate the performance of MLLMs in various aspects. It encompasses ten subtasks for the perception ability and four subtasks for the cognition ability, respectively. In this paper, we repurpose the dataset and select \emph{existence} and \emph{count} subsets to measure the object-level hallucination. The \emph{position} and \emph{color} subsets are used to measure the attribute-level hallucination.
Similar to the setup of POPE, each subset is composed of ``Yes-or-No'' questions. We report the score, namely the sum of accuracy and accuracy+ following the official implementation~\cite{fu2023mme}, in which a higher score indicates better performance and fewer hallucinations.

\textbf{LLaVA-QA90}~\cite{liu2023visual} is also used to evaluate MLLMs.
Specifically, we sample 10 description-type queries that are paraphrased in various forms to instruct an MLLM to describe an image, such as ``\texttt{Describe the following image.}'' and ``\texttt{What is the photo about?}''. LLaVA-QA90 uses images from COCO~\cite{lin2014microsoft} and adopts text-only GPT-4~\cite{openai2023gpt4} to compose queries and reference answers. We discard the reference answers, directly feed the image to GPT-4V~\cite{openai2023gpt4}, and prompt it to rate the responses regarding our designed two dimensions, \ie, accuracy and detailedness.
The prompt template is available in~\cref{ssec:prompt_evaluation}.

\paragraph{Baselines.}
We choose mainstream MLLMs as our baseline models, including mPLUG-Owl~\cite{ye2023mplug}, LLaVA~\cite{liu2023visual}, MiniGPT-4~\cite{zhu2023minigpt}, and Otter~\cite{li2023otter}.
These four MLLMs follow a ``vision encoder-interface-language model'' architecture~\cite{yin2023survey} and are trained on image-text pairs. 
Specifically, LLaVA and MiniGPT-4 adopt a simple projection layer to align multimodal embeddings. mPLUG-Owl uses a Q-Former~\cite{li2023blip} to compress visual features into a fixed number of tokens, which can be concatenated with the language embeddings. Otter adopts a similar Perceiver~\cite{jaegle2021perceiver} resampler to obtain the token compression.

\paragraph{Implementation Details.}
Our pipeline is training-free and comprises three pre-trained models apart from the MLLM to be corrected. 
We choose the LLM, GPT-3.5-turbo~\cite{brown2020language}, to fulfill the subtasks of key concept extraction, question formulation, and hallucination correction. For open-set object detection, we use Grounding DINO~\cite{liu2023grounding} to extract object counting information with default detection thresholds. Moreover, we utilize BLIP-2-FlanT5$_\text{XXL}$~\cite{li2023blip} as the VQA model to answer the attribute-related questions conditioned on the input image.

For the ``Yes-or-No'' questions, we find that the instruction-following ability of some MLLMs is somewhat weak, often outputting irrelevant texts such as pure emojis or URLs.
This is an obstacle to our correction process.
Besides, some MLLMs only output a single ``Yes'' or ``No'', which also poses a challenge to the correction.
To deal with these issues, we design two simple measures: (1) we first extract keywords, \ie, ``Yes'' and ``No'' from the responses as the answers, then combine the questions with the answers into more specific claims. For example, given a question, ``\texttt{Is there a dog in the image?}'' and a model answer, ``Yes'', we compose a more specific answer as ``\texttt{Yes, there is a dog in the image.}''; (2) we additionally feed the questions to the LLM in the correction process so that the LLM can have a better grasp of the context and task requirements.

\subsection{Experimental Results}
\label{sec:result}

\begin{table*}[!t]
\small
\centering
\aboverulesep = 0.2mm 
\resizebox{0.76\textwidth}{!}{%
\begin{tabular}{llcccc|c|c}
\toprule
Setting                      & Method                     & w/Ours & Accuracy       & Precision      & Recall         & F1-Score       & Yes Rate \\ 
\midrule
\multirow{8}{*}{\textit{Random}}      & \multirow{2}{*}{LLaVA~\cite{liu2023visual}}     & \ding{55}       & 86.00          & 87.50          & 84.00          & {\ul 85.71}    & 48.00        \\  
                             &                            & \Checkmark       & \textbf{87.67} & \textbf{95.93} & 78.67          & \textbf{86.45} & 41.00        \\  
                             & \multirow{2}{*}{MiniGPT-4~\cite{zhu2023minigpt}} & \ding{55}        & 54.67          & 57.78          & 34.67          & 43.33          & 30.00        \\  
                             &                            & \Checkmark        & 85.33          & 92.06          & 77.33          & 84.06          & 42.00        \\  
                             & \multirow{2}{*}{mPLUG-Owl~\cite{ye2023mplug}} & \ding{55}        & 62.00          & 57.26          & \textbf{94.67} & 71.36          & 82.67        \\  
                             &                            & \Checkmark        & 86.33          & 93.60          & 78.00          & 85.09          & 41.67        \\  
                             & \multirow{2}{*}{Otter~\cite{li2023otter}}     & \ding{55}        & 72.33          & 66.18          & {\ul 91.33}    & 76.75          & 69.00        \\  
                             &                            & \Checkmark        & {\ul 86.67}    & {\ul 93.65}    & 78.67          & 85.51          & 42.00        \\ 
                             \midrule
\multirow{8}{*}{\textit{Popular}}     & \multirow{2}{*}{LLaVA~\cite{liu2023visual}}     & \ding{55}        & 76.67          & 72.22          & 86.67          & 78.79          & 60.00        \\  
                             &                            & \Checkmark        & 80.67          & 83.82          & 76.00          & 79.72          & 45.33        \\  
                             & \multirow{2}{*}{MiniGPT-4~\cite{zhu2023minigpt}} & \ding{55}        & 56.67          & 58.77          & 44.67          & 50.76          & 38.00        \\  
                             &                            & \Checkmark        & 82.33          & {\ul 85.40}    & 78.00          & 81.53          & 45.67        \\  
                             & \multirow{2}{*}{mPLUG-Owl~\cite{ye2023mplug}} & \ding{55}        & 57.33          & 54.20          & \textbf{94.67} & 68.93          & 87.33        \\  
                             &                            & \Checkmark        & {\ul 83.00}    & 84.14          & 81.33          & {\ul 82.71}    & 48.33        \\  
                             & \multirow{2}{*}{Otter~\cite{li2023otter}}     & \ding{55}        & 67.33          & 61.71          & {\ul 91.33}    & 73.66          & 74.00        \\  
                             &                            & \Checkmark        & \textbf{84.33} & \textbf{88.15} & 79.33          & \textbf{83.51} & 45.00        \\ 
                             \midrule
\multirow{8}{*}{\textit{Adversarial}} & \multirow{2}{*}{LLaVA~\cite{liu2023visual}}     & \ding{55}        & 73.33          & 69.02          & 84.67          & 76.05          & 61.33        \\  
                             &                            & \Checkmark        & 80.67          & 82.86          & 77.33          & 80.00          & 46.67        \\  
                             & \multirow{2}{*}{MiniGPT-4~\cite{zhu2023minigpt}} & \ding{55}        & 55.00          & 56.88          & 41.33          & 47.88          & 36.33        \\  
                             &                            & \Checkmark        & {\ul 82.33}    & {\ul 83.92}    & 80.00          & {\ul 81.91}    & 47.67        \\  
                             & \multirow{2}{*}{mPLUG-Owl~\cite{ye2023mplug}} & \ding{55}        & 56.33          & 53.51          & \textbf{96.67} & 68.88          & 90.33        \\  
                             &                            & \Checkmark        & 81.00          & 82.07          & 79.33          & 80.68          & 48.33        \\  
                             & \multirow{2}{*}{Otter~\cite{li2023otter}}     & \ding{55}        & 66.67          & 61.16          & {\ul 91.33}    & 73.26          & 74.67        \\  
                             &                            & \Checkmark        & \textbf{83.00} & \textbf{85.61} & 79.33          & \textbf{82.35} & 46.33        \\ 
                             \bottomrule
\end{tabular}%
}
\caption{Results on POPE. w/Ours denotes MLLM responses corrected by our proposed Woodpecker. The best and second-to-best performances within each setting are \textbf{bolded} and {\ul underlined}, respectively.}
\label{tab:pope}
\end{table*}

\paragraph{Results on POPE.}
The results on POPE under the random, popular, and adversarial settings are summarized in~\cref{tab:pope}.
It can be seen that, in the random setting, MiniGPT-4 is relatively weak in perception capabilities, specifically in judging the existence of objects. 
The f1-score for MiniGPT-4 is only 43.33\%, while other baselines are all over 70\%.
In addition, mPLUG-Owl and Otter tend to be overconfident, as reflected by a high Yes Rate. 
Meanwhile, the high recall and the low precision result in a relatively low f1-score.
For all of the baselines, Woodpecker achieves consistent gains in most metrics, which indicates that our method has the ability to effectively correct object-level hallucinations.
Specifically, Woodpecker obtains a relative gain of 30.66\% for MiniGPT-4 and 24.33\% for mPLUG-Owl in terms of accuracy.

In the more challenging popular and adversarial settings, MLLMs show performance degradation to different extents, more prominent in relatively stronger baselines, such as LLaVA. Specifically, compared with the random setting, LLaVA shows a 9.33\% and 12.67\% accuracy degradation in the popular and the adversarial settings, respectively. 
This tendency suggests that MLLMs may incorrectly fit some data characteristics in the training corpus. For example, the decline in the popular setting may stem from the long-tailed data distribution~\cite{li2023evaluating}.
In contrast, equipped with a robust expert vision model, our correction method shows strong stability, making obvious improvements in various metrics for the baselines, where all accuracies exceed 80\%.
Particularly, our Woodpecker largely boosts the accuracy of mPLUG-Owl from 56.33\% to 81\% in the adversarial setting.

\begin{table}[t]
\renewcommand{\arraystretch}{1.5}
\centering
\resizebox{1.0 \columnwidth}{!}{%
\aboverulesep = 0.2mm 
\begin{tabular}{lcccccc}
\toprule
\multirow{2}{*}{Method} &
  \multicolumn{1}{c}{\multirow{2}{*}{w/Ours}} &
  \multicolumn{2}{c}{\textbf{Object-level}} &
  \multicolumn{2}{c}{\textbf{Attribute-level}} &
  \multirow{2}{*}{Total} \\ 
 &
  \multicolumn{1}{c}{} &
  \textit{Existence} &
  \textit{Count} &
  \textit{Position} &
  \textit{Color} &
   \\ \midrule
\multirow{2}{*}{LLaVA~\cite{liu2023visual}}     & \ding{55} & {\ul 195.00}    & 95.00           & 53.33          & 78.33           & 421.67          \\  
                           & \Checkmark & {\ul 195.00}    & 160.00          & 55.00          & {\ul 155.00}    & {\ul 565.00}    \\ \hline
\multirow{2}{*}{MiniGPT-4~\cite{zhu2023minigpt}} & \ding{55} & 100.00          & 61.67           & 53.33          & 65.00           & 280.00          \\  
                           & \Checkmark & 183.33          & \textbf{163.33} & {\ul 60.00}    & 121.67          & 528.33          \\ \hline
\multirow{2}{*}{mPLUG-Owl~\cite{ye2023mplug}} & \ding{55} & 101.67          & 73.33           & 58.33          & 66.67           & 300.00          \\  
                           & \Checkmark & \textbf{200.00} & 131.67          & \textbf{78.33} & 145.00          & 555.00          \\ \hline
\multirow{2}{*}{Otter~\cite{li2023otter}}     & \ding{55} & 185.00          & 95.00           & 50.00          & 118.33          & 448.33          \\  
                           & \Checkmark & 195.00          & {\ul 160.00}    & 51.67          & \textbf{165.00} & \textbf{571.67} \\ 
                           \bottomrule
\end{tabular}%
}
\caption{Results on MME. w/Ours denotes MLLM responses corrected by our proposed Woodpecker. The performance is measured by scores, where the best and second-to-best for each partition are \textbf{bolded} and {\ul underlined}, respectively.}
\label{tab:mme}
\end{table}

\paragraph{Results on MME.}
Compared with POPE, the experiments on MME is more well-rounded since it covers not only object-level but also attribute-level hallucination evaluation. The corresponding results are listed in~\cref{tab:mme}. 
We can see that, for object-level evaluation, LLaVA and Otter excel in the \emph{existence} aspect, which is also verified in the POPE evaluation, while they relatively lag in answering harder \emph{count} queries. 
In this case, our correction method is particularly effective and contributes a large score gain, ranging from +65 over LLaVA to +101.66 over MiniGPT-4.
With regard to attribute-level evaluation, baseline MLLMs tend to achieve poorer results, which suggests that they are more prone to attribute-level hallucinations. 
For example, MiniGPT-4 only achieves a score of 65 in the color split, and mPLUG-Owl merely attains 66.67.
After introducing our correction framework, these MLLMs make consistent and remarkable gains, where the score of mPLUG-Owl goes up 78.33.
In contrast, the improvements in \emph{position} are relatively small, which may be caused by two factors: 
(1) the relatively weak ability of the VQA model BLIP-2 in position reasoning;
(2) LLM may not comprehend the given bounding boxes well enough to derive position relationships by itself. 

\begin{table}[!thb]
\renewcommand{\arraystretch}{1.1}
\centering
\small
\aboverulesep = 0.2mm 
\resizebox{0.9\columnwidth}{!}{%
\begin{tabular}{lccc}
\toprule
Method                     & w/Ours & Accuracy & Detailedness \\ 
\midrule
\multirow{2}{*}{LLaVA~\cite{liu2023visual}}     & \ding{55}      & 7.1      & 7.1          \\  
                           & \Checkmark      & 7.8      & 8.6          \\ \hline
\multirow{2}{*}{MiniGPT-4~\cite{zhu2023minigpt}} & \ding{55}       & 7.0      & 6.4          \\  
                           & \Checkmark       & 8.2      & 8.8          \\ \hline
\multirow{2}{*}{mPLUG-Owl~\cite{ye2023mplug}} & \ding{55}       & 5.4      & 6.4          \\ 
                           & \Checkmark       & 5.7      & 6.4          \\ \hline
\multirow{2}{*}{Otter~\cite{li2023otter}}     & \ding{55}       & 7.0      & 6.7          \\ 
                           & \Checkmark       & 8.5      & 8.8          \\ 
                           \bottomrule
\end{tabular}%
}
\caption{Results of GPT-4V-aided evaluation. The accuracy and detailedness metrics are on a scale of 10, and a higher score indicates the better performance.}
\label{tab:gpt}
\end{table}

\paragraph{Results on LLaVA-QA90.}

\begin{figure}[!t]
    \centering
    \includegraphics[width=0.85 \columnwidth]{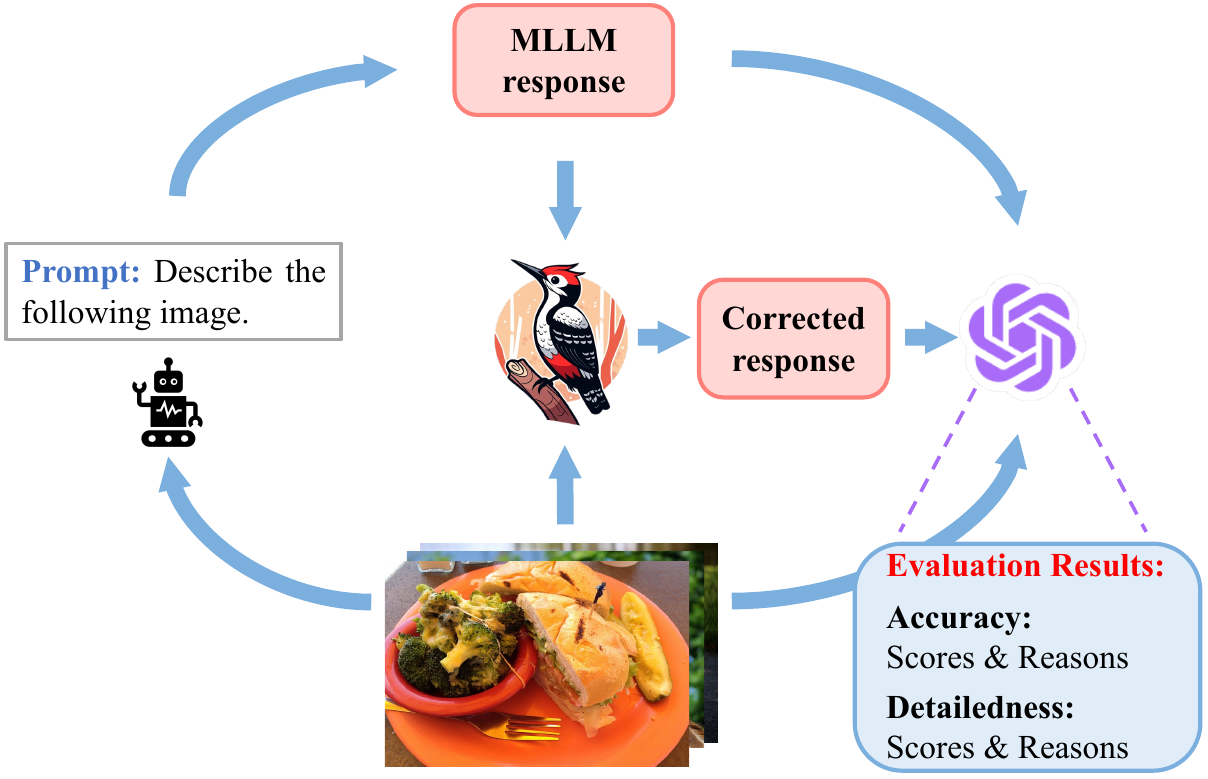}
    \caption{Illustration of GPT-4V-aided evaluation.}
    \label{fig:gpt_4v}
\end{figure}

Different from the above two experiments that only involve ``Yes-or-No'' questions, the experiment on LLaVA-QA90 is much more open. 
The description-type queries instruct MLLMs to fully translate the input image into language, rather than merely referring to the existence or the attribute of an object.

Therefore, a more reasonable and comprehensive manner is needed to support the evaluation of such open answers.
Some existing efforts are devoted to exploring automatic evaluation with the aid of LLM~\cite{liu2023mitigating,liu2023visual}. 
Specifically, a text-only GPT-4 is adopted, and the image content is fed to the language model in the form of short captions and bounding boxes of some objects. 
Nevertheless, the process of image-to-text translation inevitably loses a lot of information, making the evaluation process potentially inaccurate and biased.

In light of the recent release of a strong MLLM, GPT-4V, we propose to evaluate via a more straightforward approach.
As shown in~\cref{fig:gpt_4v}, GPT-4V can directly receive the original response, the corrected ones, and most importantly, the input image.
In such a case, we can prompt GPT-4V to let it give evaluation results and reasons for judgment.
However, it has just opened up its web interface that only supports multimodal interaction through manual operation, and there are strict limits on the number of uses.
This makes the GPT-4V-based evaluation labor-intensive, and we can only test a limited number of images, such as LLaVA-QA90.
To meet our needs, we devise the following two metrics:
\begin{itemize}
    \item Accuracy: whether the response is accurate with respect to the image content.
    \item Detailedness: whether the response is rich in details.
\end{itemize}

The scores of the two metrics are displayed in~\cref{tab:gpt}, from which we can see that our method achieves consistent gains over the baseline MLLMs. 
On the one hand, the improvement in accuracy suggests that our Woodpecker can effectively correct the hallucinations in MLLM responses. On the other hand, the bounding box information introduced in our framework adds details to the response, contributing to the boost in detailedness.

\subsection{Experimental Analysis}

\begin{figure}[!thb]
\setlength{\abovecaptionskip}{-0mm} 
\setlength{\belowcaptionskip}{-2mm} 
    \centering
    \includegraphics[width=\columnwidth]{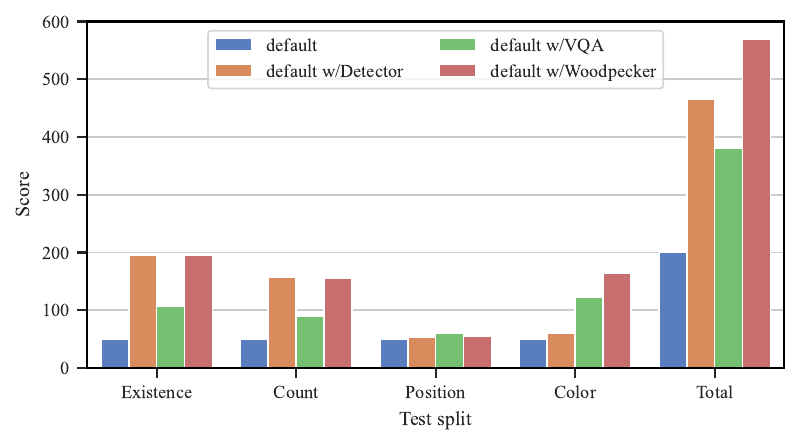}
    \caption{Results on MME with different framework variants. ``default'' is a model that always answer ``Yes'', ``default w/Detector'' introduces the object detector for hallucination correction, and ``default w/VQA'' introduces the VQA model. ``default w/Woodpecker'' is our full framework.}
    \label{fig:ablation}
\end{figure}

\paragraph{Analysis of framework modules.}
To understand the roles of different modules and their synergy, we take a dive into them and their ensemble.
For the purpose of avoiding distractions from the variation of MLLMs, we formulate a simple test bench by casting a ``default'' model that always answers ``Yes''. 
Afterward, the answers and the questions are merged into more specific claims. For example, given a question, ``\texttt{Is there a train in the picture? Please answer yes or no.}'', we compose an answer of the default model as ``\texttt{Yes, there is a train in the picture.}''. 
Furthermore, we create two extra variants of our framework, one of which only includes the open-set detector and the other with only the VQA model, respectively dubbed as ``default w/Detector'' and ``default w/VQA'':
\begin{itemize}
    \item default w/Detector. This variant is designed to probe the contribution of the detector on mitigating object-level hallucinations, more specifically, the existence and count aspects of hallucinations. 
    \item default w/VQA. By designing this variant, we aim to study the effectiveness of our selected VQA model in providing attribute information. 
\end{itemize}
The former is implemented by only providing the object-level information in the knowledge base, while the latter is realized by providing the attribute-level information. We compare these two variants with our proposed full framework, \ie, ``default Woodpecker'',  which uses both types of information.

As shown in~\cref{fig:ablation}, the gains in terms of existence and count splits mainly derive from the introduction of the open-set detector, and the improvement in the color part can be attributed to the application of the VQA model. This is in line with the expectation since we collect count information by means of the detector and gather information about specific attributes, \ie, position and color, via the VQA model. Consequently, the full model combines the advantages of both modules and achieves the best results.

To give an intuitive comprehension of the results of correction and the GPT-4V-aid evaluation, we offer a case in~\cref{ssec:exp_evaluation}. 
Specifically, we list the query and the MLLM response before and after correction. 
For reference, scores and reasons given by GPT-4V are also listed.

\begin{figure}
    \centering
    \includegraphics[width=0.8 \linewidth]{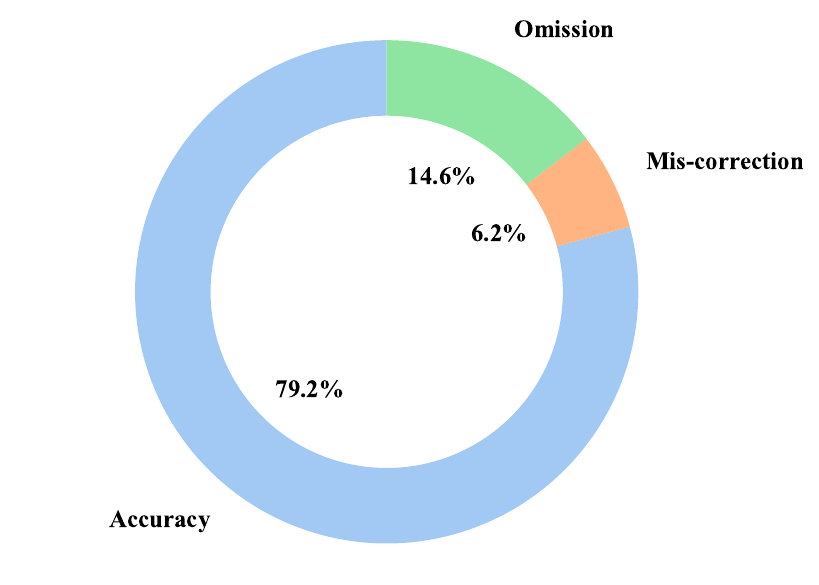}
    \caption{Proportion of different correction results.}
    \label{fig:error_analysis}
\end{figure}

\paragraph{Analysis of correction performance.}
In this part, we aim to probe further the performance of correction. 
Since there is a lack of related works in measuring the correction behavior, we fulfill this goal by breaking down the results after correction into three sections:
\begin{itemize}
    \item Accuracy: $|$ correct answers kept and wrong answers corrected $|$ $/$ $|$ problems $|$.
    \item Omission: $|$ wrong responses that fail to be corrected $|$ $/$ $|$ problems $|$.
    \item Mis-correction: $|$ correct responses mistakenly modified~$|$ $/$ $|$ problems $|$.
\end{itemize}

Concretely, we summarize the results of the ``default'' model on MME and calculate the three introduced metrics.
As reflected in~\cref{fig:error_analysis}, our correction method reaches an accuracy of 79.2\%, and meanwhile, the omission and mis-correction rates remain at a relatively low level.
The results indicate that our method can cover most cases without being over-confident.

\section{Conclusion}
\label{sec:conclusion}
In this work, we have proposed the first correction-based framework for mitigating hallucinations in MLLMs. As a training-free method, our approach incorporated multiple off-the-shelf models and could be easily integrated into different MLLMs. 
To evaluate the efficacy of the proposed framework, we conduct massive experiments on three benchmarks under different settings, including using GPT-4V for direct and automatic assessment.
We hope this work can spark new thoughts on addressing the issue of hallucinations in MLLMs.

{\small
\bibliographystyle{ieeenat_fullname}
\bibliography{11_references}
}

\ifarxiv \clearpage \appendix \section{Prompt Templates}
\label{sec:app_prompt}
In this part, we list our prompt templates for instructing LLM to fulfill various tasks, including key concept extraction, question formulation, hallucination correction, and GPT-4V-aided evaluation.

\subsection{Key Concept Extraction}
\label{ssec:prompt_entity}
The template is listed in~\cref{tab:prompt_entity}.

\begin{table*}[h!]\centering
\begin{minipage}{0.95\textwidth}
\centering
\begin{tcolorbox} 
    \centering
   
      \small
    \begin{tabular}{p{0.95\textwidth}}
   \VarSty{ {\bf System message} } \\
You are a language assistant that helps to extract information from given sentences. \\
    \midrule
   \VarSty{ {\bf Prompt} } \\
Given a sentence, extract the existent entities within the sentence for me. \\
Extract the common objects and summarize them as general categories without repetition, merge essentially similar objects. \\
Avoid extracting abstract or non-specific entities. Only extract concrete, certainly existent objects that fall in general categories and are described in a certain tone in the sentence. \\
Extract entity in the singular form. Output all the extracted types of items in one line and separate each object type with a period. If there is nothing to output, then output a single ``None''. \\
\\
Examples: \\
\textcolor[rgb]{0,0.7,0}{ \{In-context examples\} } \\
\\
Sentence: \\
\textcolor[rgb]{0.8,0,0}{\{Input sentence\}} \\
\\
Output:
    \end{tabular}
\end{tcolorbox}
\caption{Template for prompting LLM to perform key concept extraction. \textcolor[rgb]{0,0.7,0}{ \{In-context examples\} } are in-context examples used to better instruct the LLM to fulfill the task, and \textcolor[rgb]{0.8,0,0}{\{Input sentence\}} is the input from which the key concept is extracted.}
    \label{tab:prompt_entity}
\end{minipage}
\end{table*}

\subsection{Question Formulation}
\label{ssec:prompt_question}
The template is listed in~\cref{tab:prompt_question}.

\begin{table*}[h!]\centering
\begin{minipage}{0.95\textwidth}
\centering
\begin{tcolorbox} 
    \centering
   
      \small
    \begin{tabular}{p{0.95\textwidth}}
   \VarSty{ {\bf System message} } \\
You are a language assistant that helps to ask questions about a sentence. \\
    \midrule
   \VarSty{ {\bf Prompt} } \\
Given a sentence, extract the existent entities within the sentence for me. \\
Given a sentence and some entities connected by periods, you are required to ask some relevant questions about the specified entities involved in the sentence, so that the questions can help to verify the factuality of the sentence. \\
Questions may involve basic attributes such as colors and actions mentioned in the sentence. Do not ask questions involving object counts or the existence of objects. \\
When asking questions about attributes, try to ask simple questions that only involve one entity. \\ 
Ask questions that can be easily decided visually. Do not ask questions that require complex reasoning. \\
Do not ask semantically similar questions. Do not ask questions only about scenes or places. \\
Use ``where'' type questions to query the position information of the involved entities. \\
Do not ask questions about uncertain or conjecture parts of the sentence, for example, the parts described with ``maybe'' or ``likely'', etc. \\
It is no need to cover all the specified entities. If there is no question to ask, simply output a ``None''. \\
When asking questions, do not assume the claims in the description as true in advance. Only ask questions relevant to the information in the sentence. \\
Only ask questions about common, specific, and concrete entities. The entities involved in the questions are limited to the range within the given entities. \\
Output only one question in each line. For each line, first output the question, then a single ``\&'', and finally entities involved in the question, still connected by periods if multiple entities are involved. \\
\\
Examples: \\
\textcolor[rgb]{0,0.7,0}{ \{In-context examples\} } \\
\\
Sentence: \\
\textcolor[rgb]{0.8,0,0}{\{Input sentence\}} \\
\\
Entities: \\
\textcolor[rgb]{0.8,0,0}{\{Input entities\}} \\
\\
Questions:
    \end{tabular}
\end{tcolorbox}
\caption{Prompt template for question formulation. \textcolor[rgb]{0,0.7,0}{ \{In-context examples\} } are in-context examples. \textcolor[rgb]{0.8,0,0}{\{Input sentence\}} and \textcolor[rgb]{0.8,0,0}{\{Input entities\}} are the inputs, where the latter comes from the step of key concept extraction.}
    \label{tab:prompt_question}
\end{minipage}
\end{table*}

\subsection{Hallucination Correction}
\label{ssec:prompt_correction}
The template is listed in~\cref{tab:prompt_correction}.

\begin{table*}[h!]\centering
\begin{minipage}{0.95\textwidth}
\centering
\begin{tcolorbox} 
    \centering
   
      \small
    \begin{tabular}{p{0.95\textwidth}}
   \VarSty{ {\bf System message} } \\
You are a language assistant that helps to refine a passage according to instructions. \\
    \midrule
   \VarSty{ {\bf Prompt} } \\
Given a passage and some supplementary information, you are required to correct and output the refined passage in a fluent and natural style, following these rules: \\
1. The supplementary information may include some of the following parts:\\
    \hspace*{2em}``Count'' information that specifies how many instances of a certain kind of entity exist, and their associated bounding boxes; \\
    \hspace*{2em}``Specific'' information that describes attribute information specific to each entity instance, including bounding boxes, colors, etc. The information is arranged in the form of ``entity 1: [bbox]'' info of this entity. Note that the entity in ``Specific'' information corresponds to that in the ``Count'' information. \\
    \hspace*{2em}``Overall'' information that may involve information about multiple entity objects.  \\
2. Try to retain the original sentence with minimal changes. \\
3. The number of entitie instances should match the number in the ``Count'' information. Also correct the number counts if the number stated in the original sentence does not match the counting information. \\
4. If the original sentence is already correct, then just keep it. If you need to rewrite the original sentence, when rewriting, try to modify the original sentence as little as possible based on the original sentence, and use the supplementary information as guidance to correct or enrich the original sentence.  \\
5. In the refined passage, when describing entities mentioned in the ``Specific'' supplementary information, add their associated bounding boxes in parentheses right after them, in the form of ``entity([bbox])''. If multiple entities of the same kind are mentioned, then separate the box with ``;'', in the form of ``entity([bbox1];[bbox2])'' \\
\\
Examples: \\
\textcolor[rgb]{0,0.7,0}{ \{In-context examples\} } \\
------------------- \\
Supplementary information: \\
\textcolor[rgb]{0.8,0,0}{\{Input information\}} \\
\\
Passage: \\
\textcolor[rgb]{0.8,0,0}{\{Input passage\}} \\
\\
Refined passage:
    \end{tabular}
\end{tcolorbox}
\caption{Prompt template for hallucination correction. \textcolor[rgb]{0,0.7,0}{ \{In-context examples\} } are in-context examples. \textcolor[rgb]{0.8,0,0}{\{Input information\}} is the formatted knowledge base, and \textcolor[rgb]{0.8,0,0}{\{Input passage\}} is the original response to be corrected.}
    \label{tab:prompt_correction}
\end{minipage}
\end{table*}

\subsection{GPT-4V-aided Evaluation}
\label{ssec:prompt_evaluation}
The template is listed in~\cref{tab:prompt_evaluation}.

\begin{table*}[h!]\centering
\begin{minipage}{0.95\textwidth}
\centering
\begin{tcolorbox} 
    \centering
   
      \small
    \begin{tabular}{p{0.95\textwidth}}
   \VarSty{ {\bf Prompt} } \\
You are required to score the performance of two AI assistants in describing a given image. You should pay extra attention to the hallucination, which refers to the part of descriptions that are inconsistent with the image content, such as claiming the existence of something not present in the image or describing incorrectly in terms of the counts, positions, or colors of objects in the image. Note that the descriptions may be accompanied by bounding boxes, indicating the position of objects in the image, which are represented as [x1, y1, x2, y2] with floating numbers ranging from 0 to 1. These values correspond to the top left x1, top left y1, bottom right x2, and bottom right y2. \\
Please rate the responses of the assistants on a scale of 1 to 10, where a higher score indicates better performance, according to the following criteria:\\
1: Accuracy: whether the response is accurate with respect to the image content. Responses with fewer hallucinations should be given higher scores.\\
2: Detailedness: whether the response is rich in necessary details. Note that hallucinated descriptions should not count as necessary details.\\
Please output a single line for each criterion, containing only two values indicating the scores for Assistant 1 and 2, respectively. The two scores are separated by a space. Following the scores, please provide an explanation of your evaluation, avoiding any potential bias and ensuring that the order in which the responses were presented does not affect your judgment.\\
\\
\\
\lbrack{}Assistant 1\rbrack{}\\
\textcolor[rgb]{0,0.7,0}{ \{Response 1\} } \\
\\
\lbrack{}End of Assistant 1\rbrack{} \\
\\
\lbrack{}Assistant 2\rbrack{} \\
\textcolor[rgb]{0,0.7,0}{ \{Response 2\} } \\
\\
\lbrack{}End of Assistant 2\rbrack{} \\
\\
Output format:\\
\\
Accuracy:\\
Scores of the two answers:\\
Reason:\\
\\
Detailedness:\\
Scores of the two answers:\\
Reason:\\
    \end{tabular}
\end{tcolorbox}
\caption{Prompt template for GPT-4V-aided evaluation. \textcolor[rgb]{0,0.7,0}{ \{Response 1\} } and \textcolor[rgb]{0,0.7,0}{ \{Response 2\} } are the original responses and the corrected ones, respectively.}
    \label{tab:prompt_evaluation}
\end{minipage}
\end{table*}

\section{GPT-4V-aided Evaluation Case}
\label{ssec:exp_evaluation}

\begin{figure*}[!thb]
    \centering
    \includegraphics[width=1\textwidth]{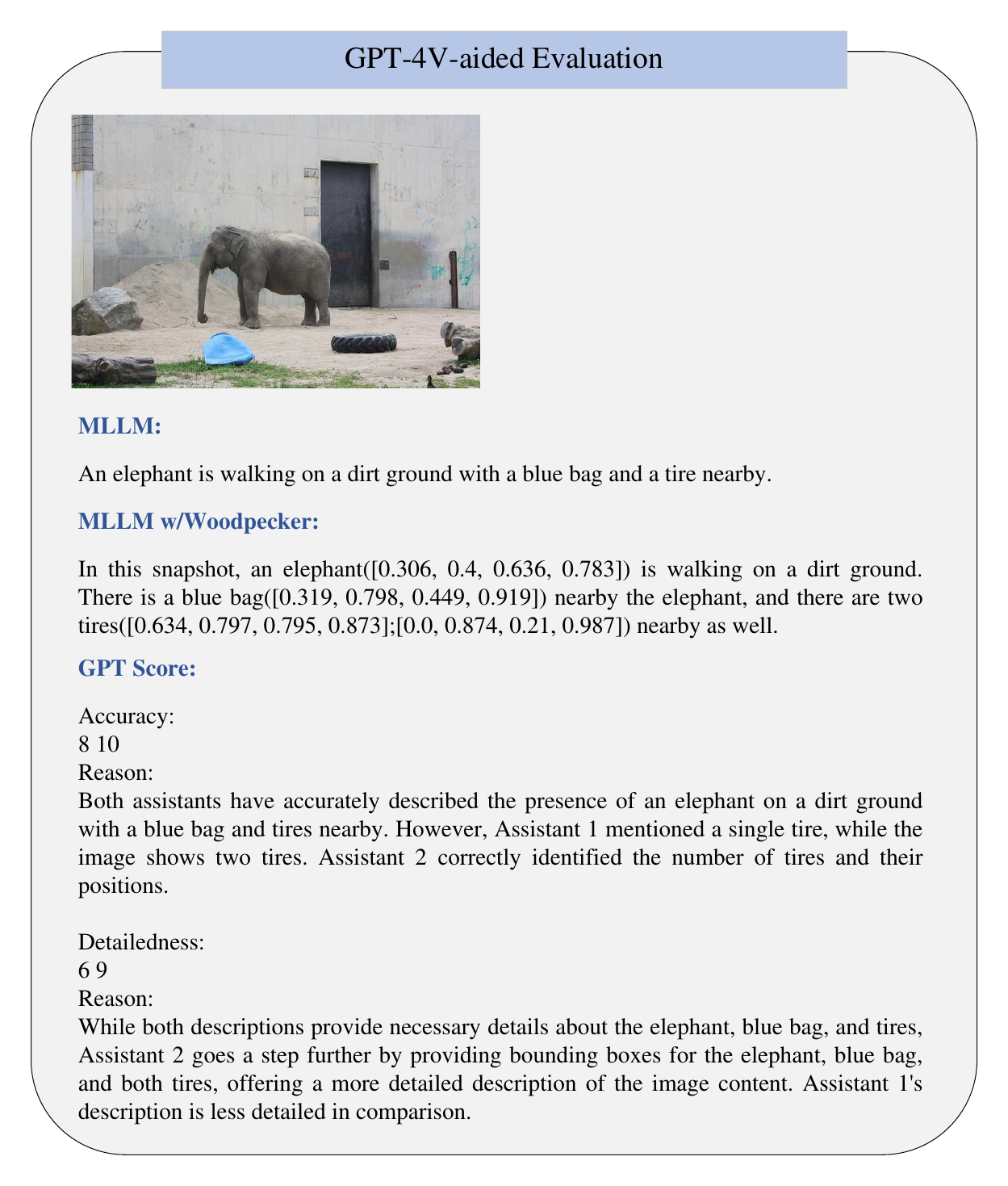}
    \caption{Example for the GPT-4V-aided evaluation.}
    \label{fig:eval_2}
\end{figure*}

To offer a straightforward and intuitive understanding, we list an evaluation case in~\cref{fig:eval_2}, where ``Assistant 1'' and ``Assistant 2'' in the evaluation reason part correspond to ``MLLM'' and ``MLLM w/Woodpecker'', respectively. GPT-4V gives not only respective scores for responses but also reasons for the judgment.

 \fi

\end{document}